\pdfoutput=1

\documentclass{nle}

\usepackage{amsmath}
\usepackage{graphicx}

\title[Using NLP to measure democracy]
      {Using NLP to measure democracy\footnote{This work was supported by the Fulbright (grantee ID 15101786); by the Coordena\c{c}\~ao de Aperfei\c{c}oamento de Pessoal de N\'ivel Superior - CAPES (BEX 2821/09-5); by the Minist\'erio do Planejamento, Or\c{c}amento e Gest\~ao - MPOG (proceed n. 03080.000769/2010-53); and by an allocation of computing time from the Ohio State Supercomputer. I thank Irfan Nooruddin, Sarah Brooks, Marcus Kurtz, Janet Box-Steffensmeier, Philipp Rehm, Paul DeBell, Carolyn Morgan, Peter Tunkis, Vittorio Merola, and Margaret Hanson for helpful comments. All errors are mine.}}
\author[Thiago Marzag\~ao]
       {T\ls h\ls i\ls a\ls g\ls o\ns M\ls a\ls r\ls z\ls a\ls g\ls \~a\ls o\ns \\
       Conselho Administrativo de Defesa Econ\^{o}mica \\
       thiago.marzagao@cade.gov.br}

\pagerange{\pageref{firstpage}--\pageref{lastpage}}
\pubyear{}

\begin{document}

\label{firstpage}
\maketitle

\begin{abstract}

This paper uses natural language processing to create the first machine-coded democracy index, which I call Automated Democracy Scores (ADS). The ADS are based on 42 million news articles from 6,043 different sources and cover all independent countries in the 1993-2012 period. Unlike the democracy indices we have today the ADS are replicable and have standard errors small enough to actually distinguish between cases.

The ADS are produced with supervised learning. Three approaches are tried: a) a combination of Latent Semantic Analysis and tree-based regression methods; b) a combination of Latent Dirichlet Allocation and tree-based regression methods; and c) the Wordscores algorithm. The Wordscores algorithm outperforms the alternatives, so it is the one on which the ADS are based.

There is a web application where anyone can change the training set and see how the results change: democracy-scores.org

\end{abstract}

\section{Introduction}

Democracy is a central variable in economics and in political science. Why are some countries more democratic than others? Why do coups happen? Does democracy impact economic policy? Does democratization affect the probability of a country going to war? These are some of the questions with which economists and political scientists concern themselves.

To answer those questions researchers need democracy to be measured somehow. That is where democracy indices come in. There are at least twelve democracy indices today (Pemstein, Meserve, and Melton 2010), the most popular of which is the Polity (Marshall, Gurr, and Jaggers 2013), which assigns a score from -10 to +10 to each of 167 countries in each year of the 1946-2013 period. Another popular democracy index is the Freedom House one (Freedom House 2013), which assigns a score from 1 to 7 to each of 195 countries in each year of the 1972-2013 period.

All democracy indices draw to some extent from Dahl's (1972) conceptualization: democracy as a mixture of competition and participation. Yet none are replicable or provide adequate measures of uncertainty. All democracy indices we have today rely directly or indirectly on country experts checking boxes on questionnaires. We do not observe what boxes they are checking, or why; all we observe are the final scores. The process is opaque and at odds with the increasingly demanding standards of openness and replicability of the field. More importantly, opacity makes it easy for country experts to boost the scores of countries that adopt the `correct' policies.

Coding rules help, but still leave too much open for interpretation. For instance, consider this excerpt from the Polity handbook (Marshall, Gurr, and Jaggers 2013): `If the regime bans all major rival parties but allows minor political parties to operate, it is coded here. However, these parties must have some degree of autonomy from the ruling party/faction and must represent a moderate ideological/philosophical, although not political, challenge to the incumbent regime.' (73). How do we measure autonomy? Can we always observe it? What is `moderate'? Clearly it is not hard to smuggle ideological contraband into democracy scores.

Ideological biases, in turn, make empirical tests circular. If we find an association between democracy and some policy $x$ is that a genuine association or an artifact of human coders' preferences regarding $x$? With the democracy measures we have today it is hard to know. Every time we regress the Polity scores or the Freedom House (Freedom House 2013) scores on some policy $x$ we may be regressing not $y$ on $x$ but $y$ on $f(x, y)$ instead.

Another problem with existing indices is the lack of proper standard errors. The two most popular indices - the Polity and the Freedom House - only give us point estimates, without any measure of uncertainty. That prevents us from knowing, say, whether Uruguay (Polity score = 10) is really more democratic than Argentina (Polity score = 8) or whether the uncertainty of the measurement process is sufficient to make them statistically indistinguishable. In other words, we cannot do descriptive inference.

Moreover, without standard errors we cannot do causal inference when democracy is one of the regressors. As Treier and Jackman (2008) warn, `whenever democracy appears as an exploratory variable in empirical work, there is an (almost always ignored) errors-in-variables problem, potentially invalidating the substantive conclusions of these studies' (203).

Only one (publicly available) measure has standard errors: the Unified Democracy Scores (UDS), created by Pemstein et al. (2010). To produce the UDS Pemstein et al. (2010) treated democracy as a latent variable and used a multirater ordinal probit model to extract that latent variable from twelve different democracy measures (among which the Polity and the Freedom House). The UDS comes with point estimates (posterior means) and confidence intervals (posterior quantiles).

The UDS is a big improvement on all other measures, but their standard errors are too large to be useful. 70\% of the countries are all statistically indistinguishable from each other (in the year 2008 - the last year in the UDS dataset at the moment of writing); pairs as diverse (regime-wise) as Denmark and Suriname, Poland and Mali, or New Zealand and Mexico have overlapping confidence intervals.

Hence we need a new democracy index - one that is replicable and comes with standard errors small enough to actually distinguish between cases. In this paper I use natural language processing to create such measure.

\section{Methods}

The basic idea is simple. News articles on, say, North Korea or Cuba contain words like `censorship' and `repression' more often than news articles on Belgium or Australia. Hence news articles contain quantifiable regime-related information that we can use to create a democracy index.

To produce the ADS I relied on supervised learning. I tried three different approaches, compared the results, and picked the approach that worked best. More specifically, I tried: a) a combination of Latent Semantic Analysis and tree-based regression methods; b) a combination of Latent Dirichlet Allocation and tree-based regression methods; and c) the Wordscores algorithm. The Wordscores algorithm outperformed the alternatives.

In this section I explain the corpus selection and the approaches I tried.

\subsection{Corpus}

I use a total of 6,043 news sources. These are all the news sources in English available on LexisNexis Academic, which is an online repository of journalistic content. The list includes American newspapers like The New York Times, USA Today, and The Washington Post; foreign newspapers like The Guardian and The Daily Telegraph; news agencies like Reuters, Agence France Presse (English edition), and Associated Press; and online sources like blogs and TV stations' websites. 

I use LexisNexis' internal taxonomy to identify and select articles that contain regime-related news. In particular, I choose all articles with one or more of the following tags: `human rights violations' (a subtag of `crime, law enforcement and corrections'); `elections and politics' (a subtag of `government and public administration'); `human rights' (a subtag of `international relations and national security'); `human rights and civil liberties law' (a subtag of `law and legal system'); and `censorship' (a subtag of `society, social assistance and lifestyle').

LexisNexis' news database covers the period 1980-present (though actual coverage varies by news source), so in principle the ADS could cover that period as well. In practice, however, LexisNexis does not provide search codes for countries that have ceased to exist, so we cannot reliably retrieve news articles on, say, the Soviet Union or East Germany (we could search by the country's name but that yields unreliable results - think of Turkey, for instance). Hence I limit myself to the 1992-2012 interval.

That selection - i.e., regime-related news, all countries that exist today, 1992-2012 - results in a total of about 42 million articles (around 4 billion words total), which I then organize by country-year. To help reduce spurious associations I remove proper nouns (that should help prevent, for instance, `Washington' being associated with high levels of democracy just because the word appears frequently on news stories featuring a democratic country), in a probabilistic way (if all occurrences of the word are capitalized then that is probably a proper noun and therefore it is removed).

For each country-year I merge all the corresponding news articles into a single document and transform it into a term-frequency vector. I then merge all vectors together in a big term-frequency matrix.

I try two variations of the same collection of news articles. The first variation, which I call corpus A, is just as described above, with no changes. It contains about 6.3 million unique words after proper nouns are removed probabilistically. The second variation, which I call corpus B, is corpus A minus: a) the 100 most frequent words in the English language; and b) any word does not appear more than once in any of the documents. Hence corpus B $\subset$ corpus A. Corpus B has about 2.3 million unique words.

\subsection{Supervised learning}

I adopt a \emph{supervised learning} approach. In supervised learning we feed the machine a number of pre-scored cases - the training data. The machine then `learns' from the training data. In text analysis that means learning how the frequency of each word or topic varies according to the document scores. For instance, the algorithm may learn that the word `censorship' is more frequent the lower the democracy score of the document. Finally, the algorithm uses that knowledge to assign scores to all other cases - i.e., to the test data.

The period 1992-2012 gives us a total of 4,067 country-years. I choose the year 1992 for the training data and extract the corresponding scores from the Unified Democracy Scores - UDS (Pemstein et al. 2010). The UDS have data on 184 countries for the year 1992. Hence we have 184 samples in the training data and 3,883 (4,067 - 184) samples in the test data. I select the year 1992 simply because it is the first year in our dataset. I select the UDS because it is an amalgamation of several other democracy scores, which reduces measurement noise.

\subsection{Latent Semantic Analysis}

Latent Semantic Analysis (LSA) is a method for extracting topics from texts \cite{landauer98}. More concretely, LSA tells us two things: a) which words `matter' more for each topic; and b) which topics appear more in each document.

We start with a term-frequency matrix, $TF_{ij}$, where rows represent terms, columns represent documents, and each entry is the frequency of term $i$ on document $j$. In our case each column represents the set of all news articles about a given country published in a given year - for instance, France-1995 or Colombia-2003.

Next we apply the $TF-IDF$ transformation. The $TF-IDF$ of each entry is given by its term-frequency ($TF_{ij}$) multiplied by $ln(n / df_{i})$, where $n$ is the total number of documents and $df_{i}$ is the number of documents in which word $i$ appears (i.e., the word's document frequency -- $DF$; the $ln(n / df_{i})$ ratio thus gives us the inverse document frequency -- $IDF$).

What the $TF-IDF$ transformation does is increase the importance of the word the more it appears in the document but the less it appears in the whole corpus. Hence it helps us reduce the weights of inane words like `the', `of', etc and increase the weights of discriminant words (i.e., words that appear a lot but only in a few documents). For more details on $TF-IDF$ see Manning, Raghavan, and Sch\"utze (2008).

The next step is normalization. Here we have documents of widely different sizes, ranging from a few kilobytes (documents corresponding to small countries, like Andorra or San Marino, which rarely appear in the news) to 15 megabytes (documents corresponding to the United States, Russia, etc - countries that appear in the news all the time). Longer documents contain more unique words and have larger $TF$ values, which may skew the results (Manning et al. 2008). To avoid that we normalize the columns of the $TF-IDF$ matrix, transforming them into unit vectors. 

We will call our normalized $TF-IDF$ matrix $A$. Its dimensions are $m \times n$ ($m$ is the number of unique words in all documents and $n$ is the number of documents). 

Now we are finally ready to run LSA. In broad strokes, LSA is the use of a particular matrix factorization algorithm - singular value decomposition (SVD) - to decompose a term-frequency matrix or some transformation thereof (like $TF-IDF$ or normalized $TF-IDF$) and extract the word weights and topic scores. Let us break down LSA into each step.

We start by deciding how many topics we want to extract - call it $k$. There is no principled way to choose $k$, but for large corpus the rule of thumb is something between 100 and 300 (Martin and Berry 2011).

To extract the desired $k$ topics we use SVD to decompose $A$, as follows:

\begin{equation}
\underset{m \times n}A = \underset{m \times m}U \enspace \underset{m \times n}\Sigma \enspace \underset{n \times n}{V^{*}} \\
\end{equation}

where $U$ is an orthogonal matrix (i.e., $U'U=I$) whose columns are the left-singular vectors of $A$; $\Sigma$ is a diagonal matrix whose non-zero entries are the singular values of $A$; and $V^{*}$ is an orthogonal matrix whose columns are the right-singular vectors of $A$ \cite{martin11}. (Notation: throughout this paper for any matrix $M$, $M'$ is its transpose and $M^{*}$ is its conjugate transpose).

There are several algorithms for computing SVD and the one I use is the one created by Halko, Martisson, and Tropp (2011), which is designed to handle large matrices.

Once we have decomposed $A$ we truncate $U$, $\Sigma$, and $V^{*}$. We do that by keeping only the first $k$ columns of $U$, the first $k$ rows and first $k$ columns of $\Sigma$, and the first $k$ rows of $V^{*}$. Let us call these truncated matrices $\tilde{U}$, $\tilde{\Sigma}$, and $\tilde{V}^{*}$.
The truncated matrices give us what we want. $\tilde{U}$ maps words onto topics: each entry $\tilde{u}_{ij}$ gives us the weight (`salience') of word $i$ on topic $j$. For instance, if we ran LSA on a collection of medical articles the resulting $\tilde{U}$ matrix might look like this:

\newpage

\begin{table}[ht!]
  \caption{Words-to-topics mapping with LSA}
  \begin{minipage}{\textwidth}
    \begin{tabular}{ccc}
    \hline\hline
    topic \#1 & topic \#2 & topic \#3 \\
    \hline
    mellitus 0.985 & atherosclerosis 0.887 & mutation -0.867 \\
    sugar 0.887 & heart -0.803 & malignant 0.542 \\
    insulin -0.867 & hypertension 0.722 & chemotherapy 0.468 \\
    ... & ... & ... \\
    heart 0.512 & mellitus 0.313 & insulin 0.207 \\
    ... & ... & ... \\
    \hline\hline
    \end{tabular}
    \vspace{-2\baselineskip}
  \end{minipage}
  \label{table1}
\end{table}

\vspace*{1\baselineskip}

The largest word weights (in absolute values) help us see what topics underlie the set of texts - in this example, diabetes (topic \#1), heart diseases (topic \#2), and cancer (topic \#3).

Importantly, each topic contains weights for all words that appear in the entire corpus. For instance, topic \#3 contains not only cancer-related words but also all other words: insulin, mellitus, heart, etc. Hence all topics have exactly the same length ($m$). What changes is the weight each topic assigns to each word. E.g., in topic \#3 cancer-related words have the largest weights - which is why we label topic \#3 `cancer'.

In real life applications the topics are usually not so clear-cut. Stopwords (`the', `of', `did', etc) often have large weights in at least some of the topics. Also, it is common for the top 20 or 50 words to be very similar across two or more topics. Finally, in real life applications with large corpora we usually extract a few hundred topics, not just three.

The product $\tilde{\Sigma}\tilde{V}^{*}=\tilde{S}$, in turn, maps topics onto documents: each entry $\tilde{s}_{ij}$ gives us the weight (`salience') of topic $i$ on document $j$. If we ran LSA on a collection of medical articles the resulting product $\tilde{S}$ might look like this:

\newpage

\begin{table}[ht!]
  \caption{Topics-to-documents mapping with LSA}
  \begin{minipage}{\textwidth}
    \begin{tabular}{cccc}
    \hline\hline
    document \#1 & document \#2 & document \#3 & ... \\
    \hline
    diabetes 0.785 & cancer 0.991 & heart diseases 0.848 & ... \\
    heart diseases 0.438 & heart diseases 0.237 & diabetes 0.440 & ... \\
    cancer 0.128 & diabetes 0.090 & cancer 0.200 & ... \\
    \hline\hline
    \end{tabular}
    \vspace{-2\baselineskip}
  \end{minipage}
  \label{table2}
\end{table}

\vspace*{1\baselineskip}

Importantly, the topics extracted with LSA are ordered: the first topic - i.e., the first column of $\tilde{S}$ - captures more variation than the second column, the second column captures more variation than the third column, and so on. Thus if we run LSA with (say) $k=200$ and then with $k=300$, the first 200 topics will be the same in both $\tilde{S}$ matrices. In other words, each topic is independent of all topics extracted after it.

That is all there is to LSA: we use SVD to decompose $A$, truncate the resulting matrices, and extract from them the word weights and topic weights.

Here I expect that LSA will generate regime-related topics - say, `elections', `repression', and so on - and also extraneous topics. I will then use tree-based methods to create democracy scores, using all topics. Afterwards I should be able to inspect which topics are influencing the scores the most, drop the rows of $\tilde{S}$ corresponding to topics that are both extraneous and influential, and generate a new, improved set of democracy scores.

There is no clear rule for setting the number of topics. Here I set the number of topics ($k$) alternately to 50, 100, 150, to see how the results change. For corpus B I also try $k=200$ and $k=300$ (I cannot do that with corpus A because of memory limitations).

\subsection{Latent Dirichlet Allocation}

LSA, discussed in the previous section, is a flexible technique: it does not assume anything about how the words are generated. LSA is at bottom a data reduction technique; its core math - truncated SVD - works as well for mapping documents onto topics as it does for compressing image files.

But that flexibility comes at the cost of interpretability. The word weights and topic weights we get from LSA do not have a natural interpretation. We know that the word weights represent the `salience' of each word for each topic, and that the topic weights represent the `salience' of each topic on each document, but beyond that we cannot say much. `Salience' does not have a natural interpretation in LSA.

That is the motivation for Latent Dirichlet Allocation (LDA), which was created by Blei, Ng, and Jordan (2003). Whereas LSA is model-free, LDA models every aspect of the data-generating process of the texts. We lose generality (the results are only as good as the assumed model), but we gain interpretability: with LDA we also get word weights and topic weights, but they have a clear meaning (more on this later).

It is unclear under what conditions LSA or LDA tends to produce superior results (Anaya 2011). The topics extracted by LDA are generally believed to be more clear-cut than those extracted by LSA, but on the other hand they are also believed to be broader (Crain et al. 2012). Hence I try both LSA and LDA and compare the results.

Just as we did in LSA, here too we start by transforming our texts into data, i.e., into a term-frequency matrix, where each entry is the frequency of term $i$ on document $j$.

Unlike what we did in LSA here we will use the term frequencies directly, without any $TF-IDF$ or normalization (LDA models the data-generating process of term frequencies, not of $TF-IDF$ values or any other transformations).

LDA assumes the following data-generating process for each document (here I draw heavily from Blei et al. 2003). We begin by choosing the number of words in the document, $N$. We draw $N$ from a Poisson distribution: $N \sim Poisson(\xi)$. Next we create a $k$-dimensional vector, $\theta$, that contains the topic proportions in the document. For instance, if $\theta=[0.3, 0.2, 0.5]$ then 30\% of the words will be assigned to the first topic, 20\% to the second topic, and 50\% to the third topic (to continue the example in the previous section these topics may be, say, `diabetes', `heart diseases', and `cancer'). We draw $\theta$ from a Dirichlet distribution: $\theta \sim Dir(\alpha)$.

Now that we have the number of words ($N$) and the topic distribution ($\theta$) we are ready to choose each word, $w$. First we draw its topic, $z$, from the $k$ topics in $\theta$, as follows: $z \sim Multinomial(\theta)$. We then draw $w$ from $p(w|z,\beta)$, where $\beta$ is a $k \times m$ matrix whose entry $\beta_{ij}$ is the probability of word $j$ being selected if we randomly draw a word from topic $i$ ($m$ is the total number of unique words in the corpus). For instance, the word `insulin' may have probability 0.05 of being selected from the topic `diabetes' and 0.001 of being selected from the topic `heart diseases'.

There are three levels in the model: $\alpha$ and $\beta$ are the same for all documents, $\theta$ is specific to each document, and $z$ and $w$ are specific to each word in each document. We disregard $\xi$ because, as Blei et al. (2003) note, $N$ is an ancillary variable, independent of everything else in the model, so we can ignore its randomness.

That is all there is to the data-generating model behind LDA. To estimate the model we need to find the $\alpha$ and $\beta$ that maximize the probability of observing the $w$s:

\begin{equation}
p(\boldsymbol{w}|\alpha,\beta)=\dfrac{\Gamma(\Sigma_{i}\alpha_{i})}{\prod_{i}\Gamma(\alpha_{i})}\displaystyle\int\left(\underset{i=1}{\overset{k}\prod}\theta_{i}^{\alpha_{i}-1}\right)\left(\underset{n=1}{\overset{N}\prod}\underset{i=1}{\overset{k}\sum}\underset{j=1}{\overset{V}\prod}(\theta_{i}\beta_{ij})^{w_{n}^{j}}\right)
\end{equation}

That function is intractable, which precludes exact inference. We need to use some approximative algorithm. I use the one created by Hoffman, Blei, and Bach \shortcite{hoffman10}, which is suitable for large matrices.

Like LSA, LDA also yields an $m \times k$ matrix of wordsXtopics weights and a $k \times n$ matrix of topicsXdocuments weights. Unlike LSA though, here these estimates have a natural interpretation. Each word weight is the probability of word $i$ being selected if we randomly draw a word from topic $j$. And each topic weight is the proportion of words in document $j$ drawn from topic $i$.

As with LSA here too I try different numbers of topics: 50, 100, and 150 for both corpora (A and B) and 200 and 300 for corpus B only.

\subsection{Tree-based regression techniques}

In this section I motivate the use of tree-based techniques and explain each in detail.

\subsubsection{Motivation}

LSA and LDA give us word weights, which map words onto topics, and topic weights, which map topics onto documents. But how do we go from that to democracy scores?

In principle we could use OLS. As explained before, our training samples are the 184 independent countries in the year 1992, whose scores we extract from the UDS, and our test samples are the 3,883 independent countries in the years 1993-2012. Thus we could, in principle: a) regress the topic scores of the 184 training samples on their respective UDS scores; and b) use the estimated coefficients to compute (`predict') the democracy scores of the 3,883 test samples.

But that would not work here. We have between 50 and 300 topics and only 184 reference cases, so with OLS we would quickly run out of degrees of freedom. Even with only 50 topics we would still be violating the `10 observations per variable' rule of thumb. And there may be all sorts of interactions and other non-linearities and to model these we would need additional terms, which would require even more degrees of freedom.

Thus I use tree-based regression techniques instead. These techniques split the observations recursively, until they are all allocated in homogeneous `leaves'. Each split - and thus the path to each leaf - is based on certain values of the regressors (if $x_{1}>10$ then follow this branch, if $x_{1} \leq 10$ then follow this other branch, etc). Tree-based algorithms are non-parametric: we are not estimating any parameters, we are simply trying to find the best splitting points (i.e., the splitting points that make the leaves as homogeneous as possible). 

Because tree-based algorithms are non-parametric we can use them even if we have more variables than observations (Gr\"{o}mping 2009). Moreover, tree-based algorithms handle non-linearities well: the inter-relations between variables are captured in the very hierarchical structure of the tree. We do not need to specify \emph{a priori} what variables interact or in what ways. Hence my choice of tree-based regression techniques over alternatives like LASSO, Ridge, or forward stepwise regression, all of which can handle a large number of variables but at the cost of ignoring non-linearities.

There are several tree-based methods and instead of choosing a particular one I try the four most prominent ones and compare the results. The next subsections explain each method in turn: decision trees, random forests, extreme random forests, and AdaBoost. In what follows I draw heavily from Hastie, Tibshirani, and Friedman (2008).

\subsubsection{Decision trees}

Decision trees were first introduced by Breiman et al. (1984). Say that we have a dependent variable $y$, $x$ independent variables, and $n$ observations, and that all variables are continuous. We want to split the observations in two subsets (let us call them $r1$ and $r2$) based on some independent variable $j$ and some value $s$. More specifically, we want to put all observations for which $j \leq s$ in one subset and all observations for which $j>s$ in another subset. But we do not want to choose $j$ and $s$ arbitrarily: we want to choose the $j$ and $s$ that make each subset as homogeneous as possible when it comes to $y$. To do that we find the $j$ and $s$ that minimize the sum of the mean squared errors of the subsets:

\begin{equation}
\underset{j,s}{\min}\left[\dfrac{\sum (\underset{y_{i} \in r1}{y_{i}} - \underset{\bar{y} \in r1}{\bar{y}})^2}{n_{r1}} + \dfrac{\sum (\underset{y_{i} \in r2}{y_{i}} - \underset{\bar{y} \in r2}{\bar{y}})^2}{n_{r2}}\right]
\end{equation}

We find $j$ and $s$ iteratively. We then repeat the operation for the resulting subsets, partitioning each in two, and we keep doing so recursively until the subsets have fewer than $l$ observations. The lower the $l$ the better the model fits the training data but the worse it generalizes to the test data. There is no rigorous way to choose $l$. Here I just follow two popular choices of $l$ ($l=2$ and $l=5$) and compare the results.

The outcome is a decision tree that relates the $x$ independent variables to $y$. For instance, if we tried to predict individual income based on socioeconomic variables our decision tree might look like this:

\begin{figure}[ht!]
  \includegraphics[width=0.5\textwidth]{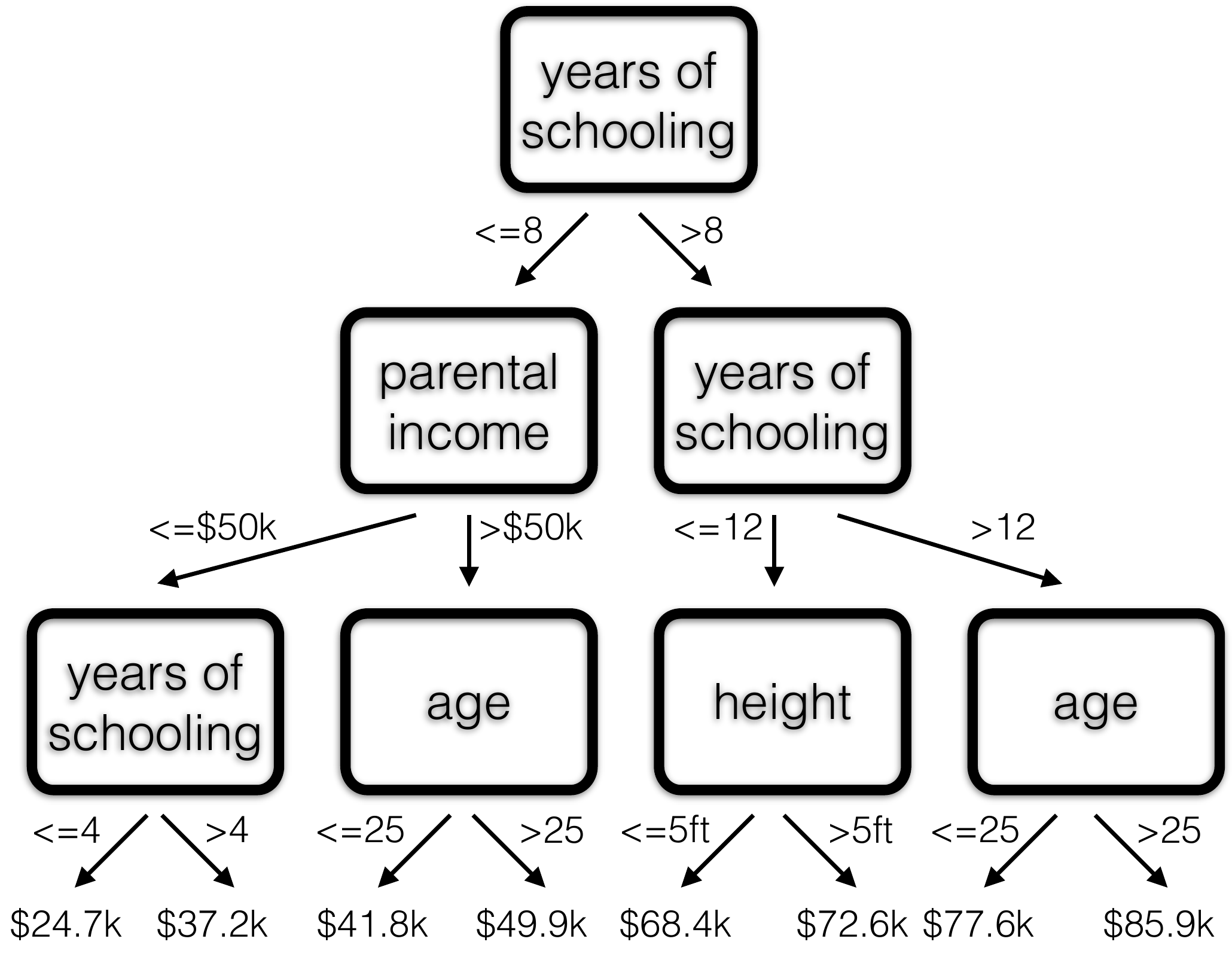}
  \caption{A (fictional) decision tree}
  \label{figure1}
\end{figure}

This is of course an extremely contrived example, but it gives us a concrete idea of what a decision tree looks like. Each node splits the observations in two groups according to some independent variable $j$ and some splitting point $s$, chosen so as to minimize the sum of the mean squared errors of the two subsets immediately under that node. We stop growing the tree when the subsets become small enough - i.e., when the subsets contain fewer than $l$ observations. The very last subsets are the leaves of the tree. (The process can be tweaked in a number of ways - we can replace the mean squared error by other criteria, we can split the subsets in more than two, we can `prune' the tree, etc -, but here I stick to the basics.)

The average $y$ of each leaf gives us the predicted $y$ for new observations. Here, for instance, someone with 5 years of schooling, parental income over \$50k, and age 30 would have a predicted income of \$49.9k.

We can see how the tree captures non-linearities. Parental income only matters when the individual has eight years of schooling or less. Height only matters when the individual has more than eight but twelve or less years of schooling. Age only matters for two groups of people: those with eight years of schooling or less and parental income over \$50k; and those with more than twelve years of schooling. And so on.

With conventional regression we would need to model all such non-linearities explicitly, positing \emph{a priori} what depends on what. And we would need several interactive and higher-order terms, which would take up degrees of freedom. With decision tree learning, however, the non-linearities are learned from the data, no matter how many or how complex they are (assuming of course that we have enough observations).

In real life applications the trees are usually much bigger, with hundreds or thousands of nodes and leaves.

\subsubsection{Random forests}

Random forests were first introduced in Breiman (2001). As the name suggests, random forests are an extension of decision trees. The idea is simple. We treat the reference set as a population, draw multiple bootstrap samples from it (with each sample having the same size as the population), and use each sample to grow a decision tree. To predict $y$ for new observations we simply average out the predicted $y$s from the different trees.

The idea of random forests is to reduce noise. With a conventional decision tree small perturbations of the data can drastically impact the choice of $j$ and $s$. By averaging out the predictions of multiple, bootstrapped trees we reduce that noise. Each bootstrapped tree yields poor predictions - slightly better than random guesses -, but their average predictions should outperform those of a conventional tree. 

There is no rigorous way to choose the number of bootstrapped trees, $N$. Here I set $N=10,000$. (I tried $N=1,000$ but the results were less stable - with the same data and parameters two different sets of of random forests with $N=1,000$ each would produce somewhat different results. With $N=10,000$ the results remain the same.)

The more the bootstrapped trees are different from each other, the more they will reduce noise in the end. Thus it is common to grow the trees in a slightly different way: instead of picking $j$ from all $x$ independent variables, we pick $j$ from a random subset of $x$, with size $c \leq x$ (common choices are $c=x/3$, $c=\sqrt{x}$ and $c=\log{x}$). The smaller the $c$, the more different the trees will be, and the more they will reduce noise in the end. But if the subset is too small (say, 1) each tree will perform so poorly that their combined performance will also be poor. Choosing the subset size is a matter of trial and error. I try $c=x/3$, $c=\sqrt{x}$, and also simply $c=x$, and compare the results.

I try $l=2$ and $l=5$ for each individual tree, just as before.

\subsubsection{Extreme random forests}

This is essentially the same as random forests, except that here we also randomize the choice of $s$. Instead of finding the best $s$ for each $j$ we draw a random $s$ for each $j$ and then pick the best $(j, s)$ combination. This usually reduces noise a bit more (at the cost of degrading the performance of each individual tree).

As with random forests, $N=10,000$; $c \in (x/3, \sqrt{x}, x)$; and $l \in (2, 5)$.

\subsubsection{AdaBoost}

AdaBoost (short for adaptive boosting) was first proposed by Freund and Schapire \shortcite{freund97}. There are several variations thereof and the one I use here is Drucker's (1997), popularly known as AdaBoost.R2, which I choose for being suitable to continuous outcomes (most AdaBoost variations are designed for categorical outcomes).

Say that we have $n$ observations. We start by choosing the number of models - in our case, trees - we want to create. We grow the first tree, $t=1$, using all observations (unlike with random forest, where we use bootstrapped samples) and weighting each observation by $w_{i}^{t}=1/n$, and compute each absolute error $|y_{i}-\hat{y}_{i}|$. We then find the largest error, $D_{t}=\underset{j=1}{\overset{n}{\max}}|y_{i}-\hat{y}_{i}|$, and use it to compute the adjusted error of every observation, $e_{i}^{t}=|y_{i}-\hat{y}_{i}|/D_{t}$. Next we calculate the adjusted error of the entire tree, $\epsilon_{t}=\underset{i=1}{\overset{n}\sum}e_{i}^{t}w_{i}^{t}$. If $\epsilon_{t} \geq 0.5$ we discard the tree and stop the process (if this happens with the very first tree that means AdaBoost failed). Otherwise we compute $\beta_{t}=\epsilon_{t}/(1-\epsilon_{t})$, update the observation weights $w_{i}^{t+1}=w_{i}^{t}\beta_{t}^{1-e_{i}^{t}}/Z_{t}$ ($Z_{t}$ is a normalizing constant), and grow the next tree, $t=2$, using the updated weights. We repeat the process until all desired trees are grown or until $\epsilon_{t} \geq 0.5$. To predict $y$ for new observations we collect the individual predictions and take their weighted median, using $\ln(1/\beta_{t})$ as weights.

Intuitively, after each tree we increase the weights of the observations with the largest errors, then use the updated weights to grow the next tree. The goal is to force the learning process to concentrate on the hardest cases. Just as in random forests, here too we end up with multiple trees and we make predictions by aggregating the predictions of all trees. Unlike in random forests, however, here the trees are not independent and we aggregate their predictions by taking a weighted median rather than a simple mean.

As with random forests, $N=10,000$ (less when $\epsilon \geq 0.5$); $c \in (x/3, \sqrt{x}, x)$; and $l \in (2, 5)$.

\subsection{Wordscores}

The Wordscores algorithm was created by Laver, Benoit, and Garry (2003) - henceforth LBG. Unlike LSA or LDA it does not need to be combined with any regression method; it is a standalone algorithm that already gives us the scores of the test samples.

We begin by computing scores for each word. Let $F_{wt}$ be the relative frequency of word $w$ on training document $t$. The probability that we are reading document $t$ given that we see word $w$ is then $P(t|w)=F_{wt} / \sum\limits_{t}F_{wt}$. We let $A_{t}$ be the \emph{a priori} score of training document $t$ and compute each word score as $S_{w}=\sum\limits_{t}(P(t|w) \cdot A_{t})$.

The second step is to use the word scores to compute the scores of the test documents (also called `virgin' documents). Let $F_{wv}$ be the relative frequency of word $w$ on virgin document $v$. The score of virgin document $v$ is then $S_{v}=\sum\limits_{w}(F_{wv} \cdot S_{w})$. To score a virgin document we simply multiply each word score by its relative frequency and sum across.

The third step is the computation of uncertainty measures for the point estimates. LBG propose the following measure of uncertainty: $\sqrt{V_{v}}/\sqrt{N^{v}}$, where $V_{v}=\sum\limits_{w}F_{wv}(S_{w}-S_{v})^{2}$ and $N^{v}$ is the total number of virgin words. The $V_{v}$ term captures the dispersion of the word scores around the score of the document. Its square root divided by the square root of $N^{v}$ gives us a standard error, which we can use to assess whether two cases are statistically different from each other.

The fourth and final step is the re-scaling of the test scores. In any given text the most frequent words are stopwords (`the', `of', `and', etc). Because stopwords have similar relative frequencies across all reference texts they will have centrist scores. That makes the scores of the virgin documents `bunch' together around the middle of the scale; their dispersion is just not in the same metric as that of the training documents.

To correct for the `bunching' of test scores LBG propose re-scaling these as follows: $S_{v}^{*}=(S_{v}-S_{\bar{v}})(\sigma_{t}/\sigma_{v})+S_{\bar{v}}$, where $S_{v}$ is the raw score of virgin document $v$, $S_{\bar{v}}$ is the average raw score of all virgin documents, $\sigma_{t}$ is the standard deviation of the training scores, and $\sigma_{v}$ is the standard deviation of the virgin scores. This transformation expands the raw virgin scores by making them have the same standard deviation as the training scores. Martin and Vanberg (2008) propose an alternative re-scaling formula, but Benoit and Laver (2008) show that the original formula is more appropriate when there are many test samples and few training samples, which is the case here.

\section{Results}

I produced a total of 235 batches of democracy scores, each covering the same country-years, i.e., all 3,883 country-years in the 1993-2012 period. The first batch uses Wordscores and corpus A. The remaining 234 batches vary by corpus (A or B), topic-extraction method (LSA or LDA), prior $\alpha$ if the topic-extraction method is LDA (symmetric $\alpha$ or asymmetric normalized $\alpha$), number of topics (50, 100, 150, 200, or 300), tree method (decision trees, random forests, extreme random forests, or AdaBoost), size of the subset of $x$ when splitting the tree nodes ($c=x/3$, $c=\sqrt{x}$, or $c=x$), and minimum node size ($l=2$ or $l=5$).

The correlation between the batch based on Wordscores and corpus A with the UDS is 0.74. Table 3 below shows the correlations between the UDS and some of the other 234 batches. The choices of corpus, $c$, $l$, and $\alpha$ did not make much difference, so Table 3 only shows the correlations obtained with corpus B, $c=x$, $l=5$, and symmetric $\alpha$. The higher the correlation, the better the performance. (Replication material and instructions can be found at thiagomarzagao.com/papers)

\begin{table}[ht!]
  \caption{Correlations with UDS}
  \begin{minipage}{\textwidth}
    \begin{tabular}{ccccc}
    \hline\hline
    & dec. tree & rand. forest & ext. rand. forest & AdaBoost \\
    \hline
    \multicolumn{5}{c}{with 50 topics} \\
    LSA & -0.13 & -0.25 & -0.23 & -0.24 \\
    LDA & -0.10 & -0.25 & -0.20 & 0.17 \\
    \hline
    \multicolumn{5}{c}{with 100 topics} \\
    LSA & -0.0009 & -0.18 & -0.18 & -0.15 \\
    LDA & 0.12 & -0.13 & -0.05 & -0.09 \\
    \hline
    \multicolumn{5}{c}{with 150 topics} \\    
    LSA & -0.07 & -0.21 & -0.16 & -0.17 \\
    LDA & -0.14 & -0.25 & -0.14 & -0.27 \\
    \hline
    \multicolumn{5}{c}{with 200 topics} \\    
    LSA & 0.01 & -0.14 & -0.14 & -0.13 \\
    LDA & -0.04 & -0.09 & 0.05 & 0.007 \\
    \hline
    \multicolumn{5}{c}{with 300 topics} \\    
    LSA & -0.02 & -0.11 & -0.10 & -0.09 \\
    LDA & 0.007 & -0.21 & -0.12 & -0.20 \\    
    \hline\hline
    \end{tabular}
    \vspace{-2\baselineskip}
  \end{minipage}
  \label{table3}
\end{table}

\vspace*{1\baselineskip}

As we observe, LSA and LDA perform poorly, with the correlations usually in the 0.10-0.20 range - way below the correlation of 0.74 obtained with Wordscores. The highest correlation obtained with LSA and LDA is 0.27 (in absolute value), which still implies an unacceptably high noise-to-signal ratio.

I inspected every topic of every LSA and LDA specification. Either the words are disparate and do not form a coherent topic or the topic is too broad or not regime-related. My initial idea was to inspect the most influential topics so I could know exactly what aspects of democracy are driving the results - and drop extraneous topics if necessary. But that is not feasible: the topics do not correspond to aspects of democracy. In a sense, all topics are extraneous. Hence the democracy scores produced with LSA and LDA are all but noise, which is why they correlate so weakly with the UDS.

The Wordscores results are clearly the best, so I inspect them more closely in the next section. I call the Wordscores results Automated Democracy Scores (ADS).

\section{Overview of the ADS}

The full 1993-2012 dataset is available for download (https://s3.amazonaws.com/thiagomarzagao/ADS.csv). Figure 2 below gives an idea of the ADS distribution in 2012.

\begin{figure}[ht!]
  \includegraphics[width=1.0\textwidth]{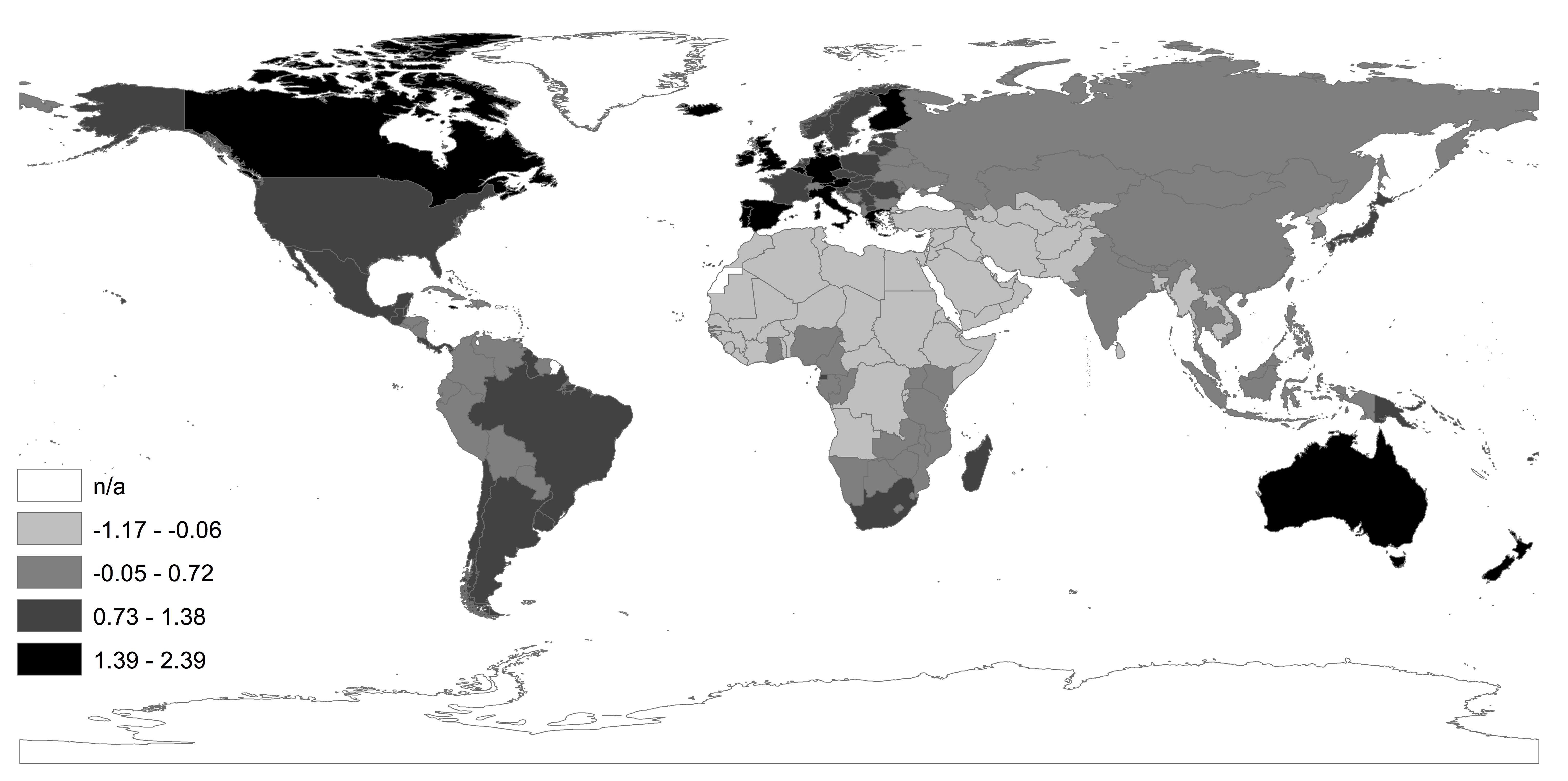}
  \caption{Automated Democracy Scores, 2012}
  \begin{footnotesize}(range limits are Jenks natural breaks)\end{footnotesize}
  \label{figure2}
\end{figure}

As expected, democracy is highest in Western Europe and in the developed portion of the English-speaking world, and lowest in Africa and in the Middle East.

Figure 3 below shows that the ADS follow a normal distribution.

\begin{figure}[ht!]
  \includegraphics[width=1.0\textwidth]{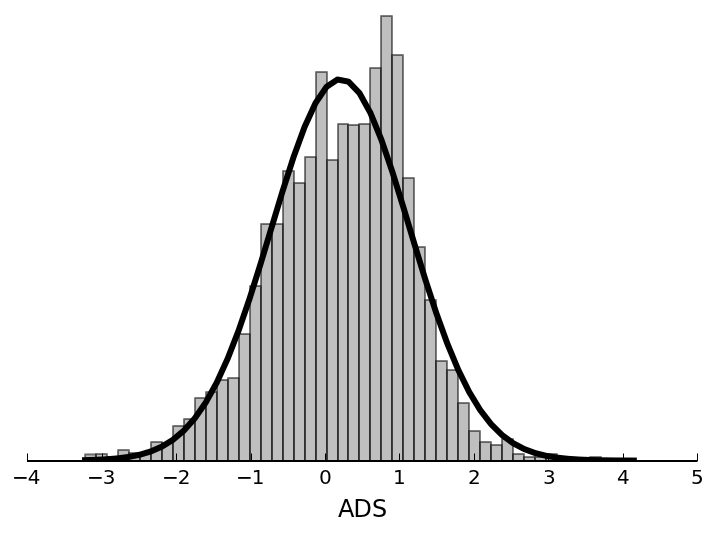}
  \caption{Automated Democracy Scores, 1993-2012}
  \begin{footnotesize}(with normal distribution)\end{footnotesize}
  \label{figure3}
\end{figure}
\newpage

Table 4 below shows the ADS summary statistics by year.

\newpage

\begin{table}[ht!]
  \caption{ADS summary statistics, by year}
  \begin{minipage}{\textwidth}
    \begin{tabular}{cccccc}
    \hline\hline
     & N & mean & std. dev. & min. & max. \\
    \hline
    1993 &193 & 0.0061666 & 1.40437 & -3.20916 & 3.81217 \\
    1994 & 193 & 0.0939503 & 1.36697 & -2.6985 & 4.14979 \\
    1995 & 193 & 0.1005004 & 1.073329 & -2.99738 & 2.36396 \\
    1996 & 193 & -0.1076104 & 1.128553 & -3.22593 & 2.26484 \\
    1997 & 193 & -0.0159435 & 1.25768 & -2.93822 & 3.03361 \\
    1998 & 193 & 0.0088406 & 1.150099 & -2.54625 & 2.7043 \\
    1999 & 193 & -0.0999732 & 1.134464 & -2.9453 & 2.63257 \\
    2000 & 193 & 0.2312175 & 0.7445582 & -1.31987 & 2.66054 \\
    2001 & 193 & 0.2222522 & 0.7182253 & -1.29777 & 1.92263 \\
    2002 & 194 & 0.2400814 & 0.735135 & -1.18534 & 2.33285 \\
    2003 & 194 & 0.2121506 & 0.7185639 & -1.3477 & 2.50623 \\
    2004 & 194 & 0.2213473 & 0.645 & -1.69878 & 2.03608 \\
    2005 & 194 & 0.3315942 & 0.6461306 & -1.08297 & 2.19639 \\
    2006 & 195 & 0.2869473 & 0.6760403 & -1.28804 & 2.18348 \\
    2007 & 195 & 0.3678394 & 0.7192703 & -1.11441 & 2.4193 \\
    2008 & 196 & 0.3860345 & 0.7002583 & -1.11659 & 2.58216 \\
    2009 & 196 & 0.3212706 & 0.6923328 & -1.487 & 2.34994 \\
    2010 & 196 & 0.4233154 & 0.6748002 & -1.08075 & 2.29522 \\
    2011 & 196 & 0.4015369 & 0.7163083 & -1.15564 & 2.38172 \\
    2012 & 196 & 0.4958635 & 0.7909505 & -1.16859 & 2.38636 \\
    all & 3883 & 0.2073097 & 0.9338698 & -3.22593 & 4.14979 \\    
    \hline\hline
    \end{tabular}
    \vspace{-2\baselineskip}
  \end{minipage}
  \label{table4}
\end{table}

\vspace*{1\baselineskip}

As expected, the average ADS increases over time, from 0.006 in 1993 to 0.495 in 2012. That reflects the several democratization processes that happened over that period. We observe the same change in other democracy indices as well (between 1993 and 2012 the average Polity score (polity2) increased from 2.24 to 4.06 and the average Freedom House score (civil liberties + political rights) decreased from 7.46 to 6.63 (Freedom House scores decrease with democracy); the average UDS score increased from 0.21 to 0.41 between 1993 and 2008, the last year in the UDS dataset).

Also as expected, the standard errors decrease with press coverage. The larger the document with the country-year's news articles, the narrower the corresponding confidence interval. As Figure 4 shows, that relationship is not linear though: after 500KB or so the confidence intervals shrink dramatically and do not change much afterwards, not even when the document has 15MB or more.

\newpage

\begin{figure}[ht!]
  \includegraphics[width=1.0\textwidth]{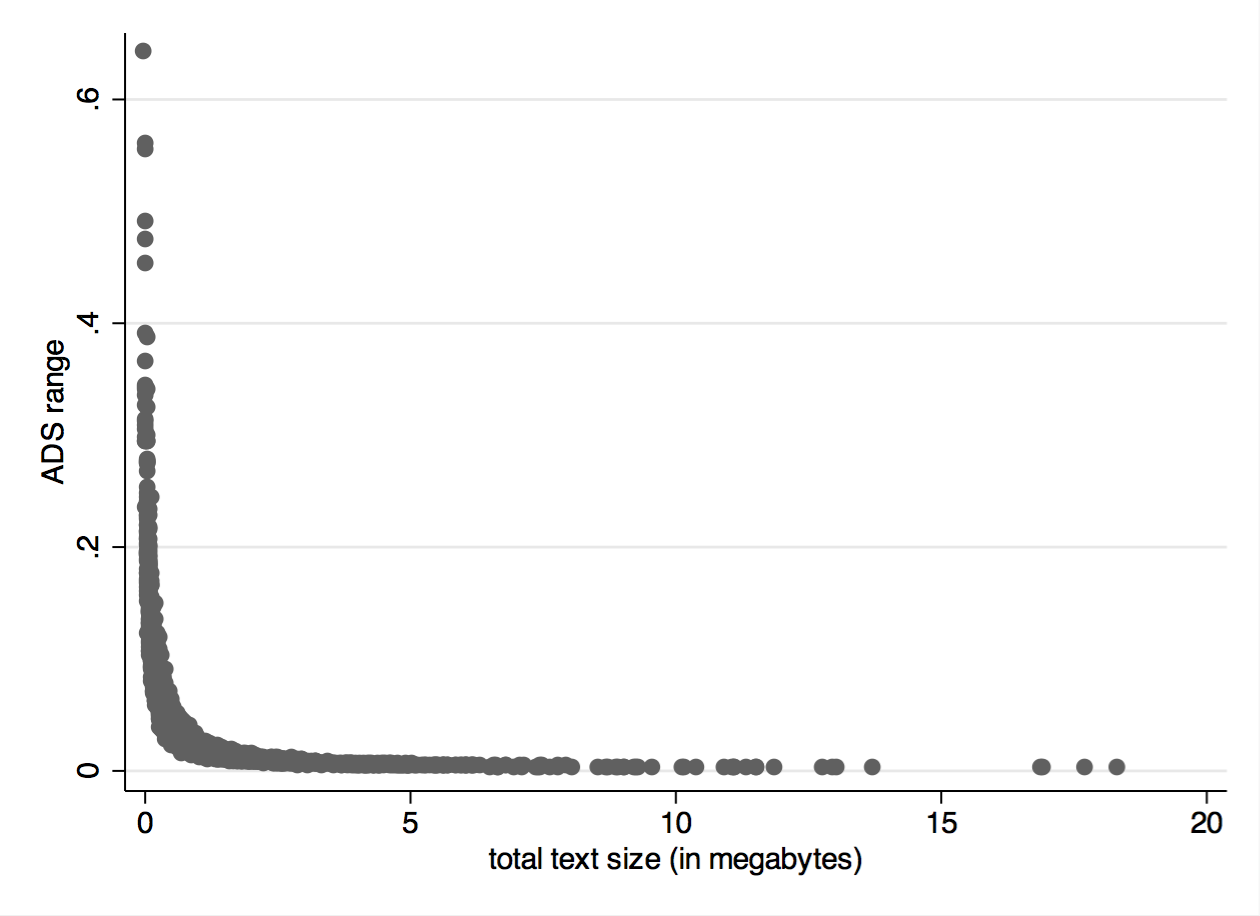}
  \caption{ADS range and press coverage}
  \begin{footnotesize}(ADS range = 95\% upper bound minus 95\% lower bound.)\end{footnotesize}
  \label{figure4}
\end{figure}

\subsection{The ADS vs other indices - point estimates}

The ADS point estimates correlate 0.7439 with the UDS' (posterior means), 0.6693 with the Polity's (polity2), and -0.7380 with the Freedom House's (civil liberties + political rights). Table 5 below breaks down these correlations by year.

\newpage

\begin{table}[ht!]
  \caption{Correlation between ADS and other indices, by year}
  \begin{minipage}{\textwidth}
    \begin{tabular}{rcccrccc}
    \hline\hline
    & UDS & Polity\footnote{polity2 (see Marshall, Gurr, and Jaggers 2013, p. 17)} & FH\footnote{civil liberties + political rights (see Freedom House 2013)} & & UDS & Polity & FH \\
    \hline
    1993 & 0.8021 & 0.7279 & -0.7677 & 2003 & 0.7470 & 0.6610 & -0.7445 \\
    1994 & 0.7921 & 0.6947 & -0.7574 & 2004 & 0.7493 & 0.6635 & -0.7553 \\  
    1995 & 0.7797 & 0.7221 & -0.7650 & 2005 & 0.7702 & 0.6833 & -0.7632 \\ 
    1996 & 0.7783 & 0.7457 & -0.7812 & 2006 & 0.7140 & 0.6458 & -0.7596 \\
    1997 & 0.8059 & 0.7647 & -0.8001 & 2007 & 0.6982 & 0.6207 & -0.7413 \\
    1998 & 0.8052 & 0.7355 & -0.7864 & 2008 & 0.7377 & 0.6363 & -0.7506 \\
    1999 & 0.7729 & 0.7260 & -0.7714 & 2009 & n/a\footnote{The UDS did not cover the 2009-2012 period at the time of writing.} & 0.6353 & -0.7627 \\
    2000 & 0.7491 & 0.6794 & -0.7579 & 2010 & n/a & 0.6467 & -0.7791 \\
    2001 & 0.7641 & 0.6881 & -0.7948 & 2011 & n/a & 0.6472 & -0.7661 \\
    2002 & 0.7668 & 0.6793 & -0.7875 & 2012 & n/a & 0.6155 & -0.7603 \\
    \hline\hline
    \end{tabular}
    \vspace{-2\baselineskip}
  \end{minipage}
  \label{table5}
\end{table}

As we see, the correlations do not vary much over time. This is a good sign: it means that the ADS are not overly influenced by the idiosyncrasies of the year 1992, from which we extract the training samples. Otherwise we would see the correlations decline sharply after 1993. The correlations do not vary much across indices either, other than being somewhat weaker for the Polity data. This is also a good sign: it means that the ADS are not overly influenced by the idiosyncrasies of the UDS, from which we extract the training scores. (Though we must remember that the UDS are partly based on the Polity and the Freedom House, so by extension the ADS also are.)

I also ran the algorithm using other years (rather than 1992) for the training data, using UDS as well. I also ran the algorithm using multiple years (up to all years but one) for the training data, again using UDS. Finally, I also ran the algorithm using not the UDS but the Polity and Freedom House indices for the training data. In all these scenarios the correlations remained in the vicinity of 0.70. This corroborates Klemmensen, Hobolt, and Hansen's (2007) finding that Wordscores' results are robust to the choice of training data.

(The samples we use for the training data cannot be used for the test data. For instance, in one scenario I used every other year for the training data, starting with 1992. In that scenario the training data was thus [1992 1994 1996 1998 2000 2002 2004 2006 2008 2010 2012] and the test data was [1993 1995 1997 1999 2001 2003 2005 2007 2009 2011]. To compute the correlations with other indices I only used the test data.) 

Country-wise, what are the most notable differences between the ADS and the UDS? Table 6 below shows the largest discrepancies.

\newpage

\begin{table}[ht!]
  \caption{Largest discrepancies between ADS and UDS}
  \begin{minipage}{\textwidth}
    \begin{tabular}{rcccrccc}
    \hline\hline
    \multicolumn{4}{c}{largest positive differences} & \multicolumn{4}{c}{largest negative differences} \\
    \hline
    & ADS & UDS & $\Delta$ & & ADS & UDS & $\Delta$ \\
    Swaziland2007 & 1.53 & -1.13 & 2.66 & Israel1994 & -1.71 & 0.97 & -2.69 \\
    Liechtenstein1994 & 4.14 & 1.56 & 2.58 & Israel1993 & -1.70 & 0.97 & -2.67 \\
    Liechtenstein1993 & 3.81 & 1.57 & 2.23 & Israel1999 & -1.20 & 1.44 & -2.65 \\
    Ireland1994 & 3.08 & 1.17 & 1.90 & Israel1997 & -1.36 & 1.07 & -2.44 \\
    Andorra1993 & 2.41 & 0.60 & 1.80 & Benin1993 & -1.79 & 0.49 & -2.28 \\
    Luxembourg1994 & 3.29 & 1.51 & 1.77 & Israel1998 & -1.17 & 1.08 & -2.26 \\
    Bhutan1996 & -0.21 & -1.97 & 1.75 & Yemen1993 & -2.60 & -0.40 & -2.20 \\
    Ireland1993 & 2.90 & 1.16 & 1.73 & Israel1996 & -1.08 & 1.08 & -2.16 \\
    Finland1994 & 3.67 & 2.00 & 1.67 & Tunisia1993 & -2.71 & -0.55 & -2.15 \\
    China2008 & 0.68 & -0.97 & 1.65 & Oman1996 & -3.18 & -1.12 & -2.05 \\
    \hline\hline
    \end{tabular}
    \vspace{-2\baselineskip}
  \end{minipage}
  \label{table6}
\end{table}

\vspace*{1\baselineskip}

The largest positive differences - i.e., the cases where the ADS are higher than the UDS - are mostly found in small countries with little press coverage. That is as expected: the less press attention, the fewer news articles we have to go by, and the harder it is to pinpoint the country's `true' democracy level. 

The largest negative differences, however, tell a different story. It seems as if either the ADS repeatedly underestimate Israel's democracy score or the UDS repeatedly overestimate it (and not only for the years shown in Table 6). We do not observe a country's `true' level of democracy, so we cannot know for sure whether the ADS or the UDS are biased (though of course these two possibilities are not mutually exclusive) but the ADS should be unbiased to the extent that we managed to filter out news articles not related to political regime; whatever biases exist in the UDS should become, by and large, random noise in the ADS.

For instance, imagine that the UDS are biased in favor of countries with generous welfare, like Sweden. The UDS of these countries will be `boosted' somewhat. But to the extent that the news articles we selected are focused on political regime and not on welfare policy, Wordscores will not associate those boosted scores with welfare-related words and hence the ADS will not be biased. The ADS will be less efficient, as (ideally) no particular words will be associated with those boosted scores, but that is it.

The UDS, on the other hand, rely on the assumption that `raters perceive democracy levels in a noisy but unbiased fashion' (Pemstein et al. 2010:10), which as Bollen and Paxton (2000) have shown is simply not true. Hence whatever biases exist in the Polity, Freedom House, etc, wind up in the UDS as well. The data-generating process behind the UDS does not mitigate bias in any way.

In other words, it seems more likely that the UDS are overestimating Israel's democracy scores than that the ADS are underestimating them. This pro-Israel bias is interesting in itself, but it also raises the more general question of whether the UDS might have an overall conservative bias. To investigate that possibility I performed a difference-of-means test, splitting the data in two groups: country-years with left-wing governments and country-years with right-wing governments (I used the EXECLRC variable from Keefer's (2002) Dataset of Political Institutions for data on government ideological orientation.)

The test rejected the null hypothesis that the mean ADS-UDS difference is the same for the two groups:  the mean ADS-UDS difference for left-wing country-years (-0.127, std. error = 0.024, n = 802) is statistically smaller than the mean ADS-UDS difference for right-wing country-years (-0.328, std. error = 0.025, n = 603), with p $<$ 0.00001. As both means are negative, it seems that the UDS tend to reward right-wing governments.

I also checked whether the UDS may be biased toward economic policy specifically. I split the country-years in the Index of Economic Freedom (Heritage Foun- dation 2014) dataset into two groups: statist (IEF score below the median) and non-statist (IEF score above the median). The difference-of-means test shows that the mean ADS-UDS difference for statists (-0.132, std. error = 0.0196, n = 1057) is statistically lower than that of non-statists (-0.215, std. error = 0.015, n = 1977), with p $<$ 0.0006. Both means are negative here as well, so it seems that the UDS somehow reward free market policies.

We cannot conclusively indict the UDS or its constituent indices though. Perhaps democracy and right-wing government are positively associated and the ADS are somehow less efficient at capturing that association. This is consistent with the Hayek-Friedman hypothesis that left-wing governments are detrimental to democracy because economic activism expands the state's coercive resources (Hayek 1944; Friedman 1962). As we do not observe a country's true level of democracy, it is hard to know for sure what is going on here.

At least until we know whether the UDS are biased or the ADS are inefficient, the ADS are the conservative choice. Say we regress economic policy on the UDS and find that more democratic countries tend to have less regulation. Is that relationship genuine or is it an artifact of the UDS being biased in favor of free market policies? With biased measures our tests become circular: we cannot know the effect of $x$ on $y$ when our measure of $x$ is partly based on $y$. Inefficiency, on the other hand, merely makes our tests more conservative.

\subsection{The ADS vs other indices - standard errors}

The ADS have much smaller standard errors than the UDS (the only other democracy index that also comes with standard errors). On average, each country in the ADS dataset in the year 2008 overlaps with other 4.49 countries; in the UDS dataset that average is 99.67. The ADS confidence intervals tend to be larger the less press coverage the country gets, but in all cases they are smaller than the corresponding UDS ones. 

For instance, in the UDS the United States is statistically indistinguishable from 80 other countries, whereas in the ADS the United States is statistically indistinguishable from only one other country (Solomon Islands, which rarely appears in the news and thus has a wide confidence interval). The country with most overlaps in the UDS data is Sao Tome and Principe, which is statistically indistinguishable from 135 other countries. That makes 70\% of the UDS scores (for 2008) statistically the same. The worst case in the ADS is Czech Republic, which overlaps with 25 other countries (in the UDS Czech Republic overlaps with 110 other countries).

The reason why the ADS standard errors are much smaller than the UDS ones is the sheer size of the data. We have 42 million news articles, which give us about 4 billion words in total. Because the total number of virgin words goes in the denominator of the formula for the standard errors, those 4 billion words shrink the confidence intervals dramatically.

The ADS standard errors also tell us something about the nature of democracy. The large standard errors of the UDS data might lead us to believe that democracy is better modeled as a categorical variable, like the one in Alvarez et al. (1996). Gugiu and Centellas (2013) claim that that is indeed the case: they use hierarchical cluster analysis to extract the latent democracy variable behind five existing indices (among which the Polity and Freedom House indices) and find that that latent variable is categorical, not continuous.

That conclusion is unwarranted though. If the constituent measures (Polity, Freedom House, etc) are too coarse to capture fine-grained regime differences then it is not surprising that their latent variable will also be too coarse to capture fine-grained regime differences. But just because a given measure fails to capture subtle distinctions does not mean that these distinctions do not exist. As the ADS standard errors suggest, these subtle distinctions do seem to exist.

\section{Conclusion}

The ADS address important limitations of the democracy indices we have today. The ADS are replicable and have standard errors narrow enough to distinguish cases. The ADS are also cost-effective: all we need are training documents and training scores, both of which already exist; there is no need to hire dozens of country experts and spend months collecting and reviewing their work.

To facilitate replicability and extensibility I created a web application where anyone can tweak the training data and see how the results change: democracy-scores.org.

It would be interesting to replicate existing (substantive) work on democracy but using the ADS instead, to see how the results change. The ADS come with standard errors, so we could incorporate these in the regressions, perhaps using errors-in-variables models (1987).

We could extend the method here to produce a daily or real-time democracy index. Existing indices are year-based and outdated by 1-12 months, so we do not know how democratic a country is \emph{today} or how democratic it was, say, on 11/16/2006. Automated text analysis can help us overcome those limitations. We cannot score the news articles from only one or two days, as there would not be enough data to produce meaningful results, but we can pick, say, the 12-month period immediately preceding a certain date - for instance, 11/17/2005-11/16/2006 if we want democracy scores for 11/16/2006.

\label{lastpage}

\end{document}